\documentclass[10pt,twocolumn,letterpaper]{article}

\usepackage{cvpr}
\usepackage{times}
\usepackage{epsfig}
\usepackage{graphicx}
\usepackage{amsmath}
\usepackage{amssymb}
\usepackage{pifont}
\usepackage{enumitem}

\usepackage{xcolor}

\newcommand{\cmark}{\ding{51}}%
\newcommand{\xmark}{\ding{55}}%
\newcommand{\hadamard}{\odot}
\newcommand\nocell[1]{\multicolumn{#1}{c|}{}}

\usepackage[pagebackref=true,breaklinks=true,letterpaper=true,colorlinks,bookmarks=false]{hyperref}

\cvprfinalcopy 

\def\cvprPaperID{****} 

\ifcvprfinal\fi

\makeatletter
\let\origparagraph\paragraph
\renewcommand\paragraph{\@ifstar{\starparagraph}{\nostarparagraph}}
\newcommand\nostarparagraph[1]
{\origparagraph{#1}\paragraphpostlude}
\newcommand\starparagraph[1]
{\paragraphprelude\origparagraph*{#1}\paragraphpostlude}
\newcommand\paragraphprelude{%
  \vspace{-15pt}
}
\newcommand\paragraphpostlude{%
}
\makeatother

\begin{document}

\title{ReConvNet: Video Object Segmentation \\ with Spatio-Temporal Features Modulation}


\author{Francesco Lattari$^1$\thanks{The authors equally contributed.},~~~
		Marco Ciccone$^1$\footnotemark[1],~~~
        Matteo Matteucci$^1$,~~~
        Jonathan Masci$^3$,~~~ 
        Francesco Visin$^{1,2}$\thanks{Most of the work has been carried out while at Politecnico di Milano} \\\\
        $^1$Politecnico di Milano,~~~
        $^2$Google DeepMind,~~~
        $^3$NNAISENSE\\
        {\tt\small \{francesco.lattari,~ 
        			 marco.ciccone,~ 
                     matteo.matteucci\}@polimi.it}, \\
        {\tt\small visin@google.com, jonathan@nnaisense.com} \\             
}
\maketitle

\begin{abstract}
We introduce \emph{ReConvNet}, a recurrent convolutional architecture 
for semi-supervised video object segmentation that is able to fast adapt 
its features to focus on any specific object of interest at inference time. 
Generalization to new objects never observed during training is 
known to be a hard task for supervised approaches that would need to be retrained.
To tackle this problem, we propose a more efficient solution that learns 
spatio-temporal features self-adapting to the object of interest via
conditional affine transformations.
This approach is simple, can be trained end-to-end and does not necessarily require extra training steps at inference time. 
Our method shows competitive results on DAVIS2016 with respect to state-of-the art
approaches that use online fine-tuning, and outperforms them on DAVIS2017.
ReConvNet shows also promising results on the DAVIS-Challenge 2018 winning the $10$-th position.
\end{abstract}

\section{Introduction}
Semi-Supervised Video Object Segmentation is the task of segmenting specific 
objects of interest in a video sequence, given their segmentation in the first frame.
This poses an interesting challenge for standard supervised methods, as the
model cannot be trained to discriminate between a fixed set of classes based on
semantics \cite{Visin2016reseg}, but rather has to learn to segment unseen objects 
based on a single instance. 
Classic supervised techniques fail to
generalize easily to new objects whose traits are potentially very different from 
the training data. This problem is known in the literature as ``domain adaptation''.

Most of the proposed approaches cast the problem as a one-shot
learning task: after a first general pre-training on the entire dataset, at
inference time the generic model is adapted into an object-specific one by
fine-tuning on transformations of the first frame for each test sequence~\cite{Caelles2017,  voigtlaender17BMVC}. 
This procedure is computationally expensive, requiring several extra steps 
of back-propagation for each sequence.
Furthermore, it heavily depends on the ability of the data augmentation procedure 
to produce realistic sequences. Indeed, generating high-quality sequences is 
clearly a very hard and ill-posed task that adds an extra layer of complexity to the problem. 

\begin{figure*}[h!]
    \centering
    \includegraphics[width=0.9\textwidth]{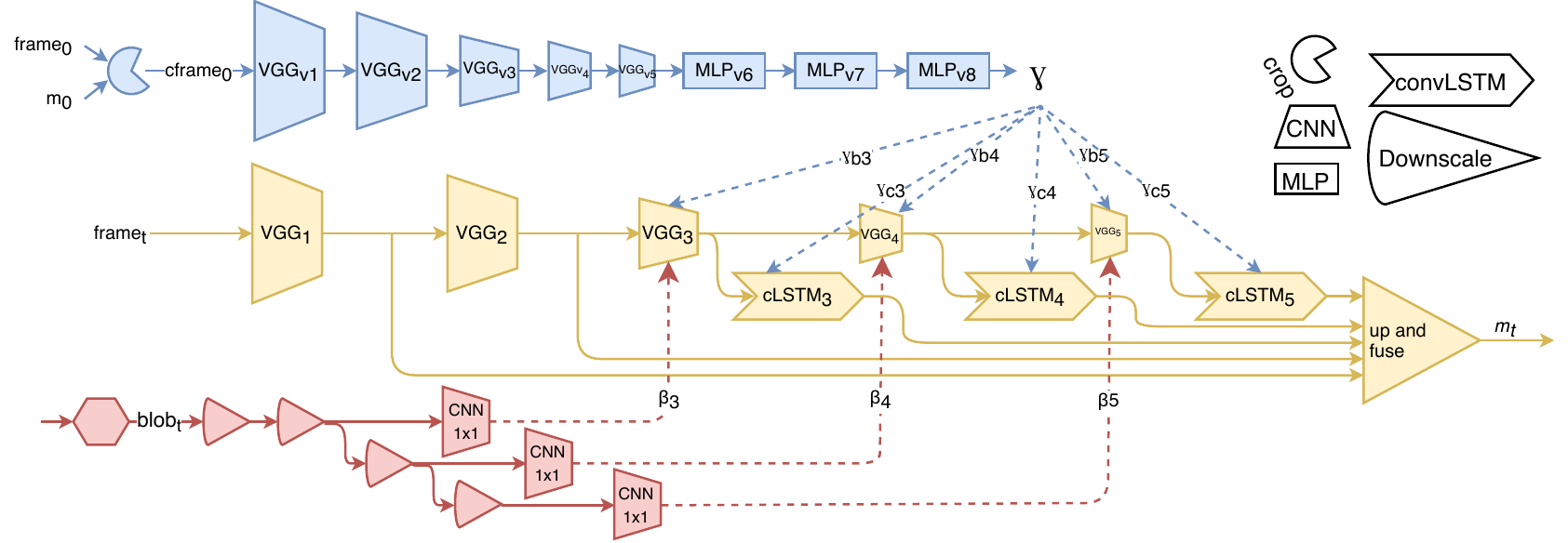}
    \vspace{-0.1cm}
    \caption{{\bf ReConvNet architecture}. The Segmentation Network (yellow) is a 
    	VGG-16-based architecture enhanced with convLSTM layers. The Visual Modulator 
        (blue) biases the SN to be selective toward the object of interest
        via a per-channel multiplicative interaction, while the Spatial Modulator (red)
        enforces a spatial prior via an additive per-location modulation.}
    \label{fig:reconvnet}
    \vspace{-0.5cm}
\end{figure*}

In this work we propose a system that can adapt to the specific object 
of interest at inference time without the need of an expensive fine-tuning 
stage. This mechanism is inspired by \emph{fast-weights}~\cite{schmidhuber1992fastweights}.
In this framework a \emph{slow network} generates, on-line, the weights of a second 
one, called \emph{fast network}, to fast adapt to a new task or to a change 
in the environment.
More recently \emph{HyperNetworks}~\cite{ha2016hypernetworks} propose to use a
network to infer a transformation of the weights rather than the weights themselves.
Similarly, \emph{FiLM} \cite{perez2017film} proposes an extension of conditional batch-normalization 
where a module produces an affine transformation that is applied to the 
features of each layer of the main neural network in order to impose a more explicit conditioning.

In the context of object segmentation, features modulation has been 
investigated on the DAVIS dataset by OSMN~\cite{Yang2018osmn}, that extends 
OSVOS~\cite{Caelles2017} by specializing a part of the architecture 
to condition the predictions on the target object.
Specifically, this is achieved by modulating the activations of the Segmentation
Network (SN) via a single forward pass of a Modulating Network (MN) that outputs a 
scale parameter for each channel based on the object features, and a shift parameter 
for each location that acts as a spatial attention mechanism similar to~\cite{dasnet2014}. 
Here the SN learns to perform generic segmentation and the MN fast-adapts it to 
attend to the object of interest.

In this paper we introduce \emph{ReConvNet}, a recurrent convolutional architecture for
semi-supervised video object segmentation. Our architecture self-adapts to segment unseen object 
without the need of extra fine-tuning steps of supervision. 
The main contributions of this work are the following:
\begin{itemize}[noitemsep]
	\item We extend the successful OSMN architecture with the possibility to model 
    	highly non-linear intrinsic temporal correlations between consecutive frames via
		convLSTM units. This modification outperforms the baseline.
    \item On DAVIS2016 we show comparable performance to methods that make use of 
    	online fine-tuning and we outperform them on the more challenging DAVIS2017.
    \item We place $10$-th in the DAVIS challenge 2018 without resorting to online fine-tuning 
    	or other post-processing steps.
   	\item We show that features modulation is orthogonal to online fine-tuning and that
    	indeed combining the two results in a further performance boost.
\end{itemize}

\section{Architecture}
ReConvNet is composed of three main components: the Segmentation Network (SN), 
the Visual Modulator (VM) and the Spatial Modulator (SM).

\paragraph*{Segmentation Network} 
The original OSMN Segmentation Network is a Fully Convolutional Network (FCN), 
based on VGG-$16$~\cite{simonyan2014very} layers that are propagated to the decoder 
after upsampling in a hyper-column fashion~\cite{hariharan2015hypercolumns} to recover details at multiple scales. 
Hence, OSMN processes each frame independently, often resulting in segmentation masks that 
lack temporal consistency and exhibit high variance across the sequence.

A natural way to incorporate temporal structure into the model is to add recurrent units.
Here we use convLSTM layers, an adaptation of the original
LSTM~\cite{hochreiter1997long} cell that takes into account the spatial
structure of its input. As proposed by~\cite{xingjian2015clstm}, the LSTM cell
can be modified by replacing the matrix multiplications of each transformation
with convolution operations. The convLSTM blocks are interleaved to the last three VGG-$16$ layers 
to endow the network with multi-scale spatio-temporal processing capability.

\paragraph*{Visual Modulator} The visual modulator is in charge of biasing the
activations of the segmentation network to target the object of interest.
This strong conditioning is achieved by a VGG-$16$ network that takes as input the
first frame cropped around the target object and resized to $224\times224$, and
produces a set $\mathbf{\Gamma}$ of vectors of \emph{scaling coefficients} --
one for each of the last three convolutional layers of the SN. In addition to the 
coefficients computed by OSMN, we also compute those for the convLSTM modules.
All the visual modulation coefficients are multiplied to the feature maps 
$f_i = \mathbf{\gamma}_i \hadamard f_i$, where $\hadamard$ indicates channel-wise multiplication. 
This has the effect of enhancing the maps related to the target object and suppressing 
the least useful, potentially distracting ones, allowing the segmentation network to quickly 
adapt to the object of interest without resorting to expensive steps of back-propagation at 
inference time. 

\paragraph*{Spatial Modulator} To help discriminating between multiple instances of
the same object and, more generally, to provide a loose prior on the location of
the target object, the network is also enriched with a spatial attention mechanism.
A rough estimate of the position of the target object can be obtained by fitting a ``gaussian
blob'' on the segmentation predicted at time $t$ and fed to a Spatial Modulator component. 
This, in turn, produces a set of \emph{shift coefficients}~$\mathbf{\beta}$, one for each of
the last three VGG layers, via a $1\times1$ convolution applied on the blob downsampled to 
the layer's resolution. Note that, as opposed to the VM case, we do not generate modulation
coefficients for the convLSTM layers. The spatial coefficients are summed pixel-wise to 
the activations of the corresponding layers, therefore shifting the focus on the parts 
of the image where the object is more likely located. The combination of VM and SM, 
$f_i = \gamma_i \hadamard f_i + \mathbf{\beta}_i$, is then applied to the features.

\section{Experiments}
\label{sec:exp}
In this section we describe the experimental settings used to evaluate the
\emph{ReConvNet} model.

\paragraph*{Experimental setting} To ensure a fair comparison between
\emph{ReConvNet} and the OSMN baseline we initialize the
components of the OSMN model in our architecture with the pretrained weights
as provided by the authors. This is done on both DAVIS2016 and DAVIS2017,
to ensure that any improvement can be 
clearly attributed to the introduction of a recurrent architecture. 

For the initialization of the remaining modules, in an effort to minimize the
factors of variations with respect to the baseline, the extra channels of the
visual modulator are initialized as in~\cite{Yang2018osmn} and, similarly, the
input-to-hidden convolutions in the convLSTM layers use the same
initialization as the convolutional layers in the baseline. Lastly, the
hidden-to-hidden convolutions are initialized to be
orthogonal~\cite{saxe2013exact}.

The model is trained with a lower learning rate
for the non-recurrent than for the recurrent component,
namely
$10^{-6}$ and $10^{-5}$ respectively. We found it beneficial to train
with the Lovasz loss~\cite{berman2017lov} that directly optimizes the
IoU measure. Finally, to prevent overfitting we employed early-stopping and data 
augmentation~\cite{visin_dataset} of the inputs of the visual and spatial modulators with 
random shift, scale, and rotation transformations.
When online fine-tuning is used, the model is trained on each test sequence with  
random transformation of the first frame for $300$ iterations and learning rate $10^{-6}$ 
for all components. 

\paragraph*{Pretraining}
In order to make use of the relative abundance of segmented static images for
pre-training, we split the training process in two phases. 
First, we train the non-recurrent components of the model on 
MSCOCO~\cite{lin2014microsoft} to learn segmentation coupled with modulation, 
then we train the full network on DAVIS2017 to account for the spatio-temporal
recurrent component
as well as to make use of the modulation to focus on the target object
throughout the sequence. 

The single frame pre-training procedure 
proved to be an essential proxy to bootstrap the temporal-consistent
segmentation. In fact, DAVIS contains only a few examples for most
semantic classes, making it very easy for the network to overfit on the
training examples.

As it can be expected, the problem is exacerbated by models with high capacity,
and even more by those that exploit a visual modulator to tackle semi-supervised
segmentation.
As shown in~\cite{Yang2018osmn} the set of parameters $\mathbf{\Gamma}$, produced by the
visual modulator, pushes the model to learn a semantic mapping in an embedding 
space where visually similar objects are close in $\ell_2$ distance. 
Learning this mapping requires a large enough amount of diverse examples.
We choose the MSCOCO~\cite{lin2014microsoft} dataset for its
wide range of classes and intra-class variations.


\begin{table*}[h!]
\small
\centering
\begin{tabular}{l|c|c|c|c|c|c|c|c|c|c|c|c|c|c|c}
\cline{3-16}
            \nocell{2}&\multicolumn{7}{|c|}{DAVIS2016} 
			&\multicolumn{7}{|c|}{DAVIS2017} \\ 
\hline
Method & FT & $\mathcal{J}$\&$\mathcal{F}$ 
			&\multicolumn{3}{|c|}{$\mathcal{J}$} 
            &\multicolumn{3}{|c|}{$\mathcal{F}$} 
            & $\mathcal{J}$\&$\mathcal{F}$
			&\multicolumn{3}{|c|}{$\mathcal{J}$} 
            &\multicolumn{3}{|c|}{$\mathcal{F}$} \\ \cline{3-16} 
       & 
       & M $\uparrow$
       & M $\uparrow$& R $\uparrow$& D $\downarrow$ 
       & M $\uparrow$& R $\uparrow$& D $\downarrow$
       & M $\uparrow$
       & M $\uparrow$& R $\uparrow$& D $\downarrow$ 
       & M $\uparrow$& R $\uparrow$& D $\downarrow$  \\
\hline
\hline
OSMN (2nd) \cite{Yang2018osmn} & \xmark & - & 74.0 & - & - & - & - & - & 54.8 & 52.5 & 60.9 & 21.5 & 57.1 & 66.1 & 24.3 \\
ReConvNet (ours) & \xmark & 78.1 & \textbf{79.4} & 89.6 & 7.7  & 76.8 & 86.6 & 7.7 & \textbf{65.7} & \textbf{62.7} & \textbf{70.5} & 21.6 & \textbf{68.7} & \textbf{77.3} & \textbf{21.6} \\
\hline
\hline
OSVOS \cite{Caelles2017} & \cmark & 80.2 & 79.8 & 93.6 & 14.9 & 80.6 & 92.6 & 15.0 & 60.3 & 56.6 & 63.8 & 26.1 & 63.9 & 73.8 & 27.0 \\
onAVOS \cite{voigtlaender17BMVC} & \cmark & 85.5 & 86.1 & 96.1 & 5.2  & 84.9 & 89.7 & 5.8 & 65.4 & 61.6 & 67.4 & 27.9 & 69.1 & 75.4 & 26.6\\
OSVOS-S \cite{Man18b} & \cmark & 86.6 & 85.6 & 96.8 & 5.5  & 87.5 & 95.9 & 8.2 & 68.0 & 64.7 & 74.2 & \textbf{15.1} & 71.3 & 80.7 & \textbf{18.5} \\
ReConvNet (ours) & \cmark & 85.0 & 85.4 & 95.9 & 8.5 & 84.6 & 93.9 & 12.1 & \textbf{70.2} & \textbf{66.6} & \textbf{75.4} & 28.1 & \textbf{73.7} & \textbf{83.1} & 29.6 \\
\hline
\hline
\end{tabular}
\caption{Comparisons of our approach vs OSMN baseline and top-3 state-of-the-art algorithms on DAVIS 2016 and 2017 validation sets. \textbf{Legend.} FT: Online fine-tuning on the first frame; M: Mean; R: Recall; D: Decay.}
\label{tab:davis2016}
\vspace{-12pt}
\end{table*}

\subsection{Single Object Segmentation}
We first evaluate our model on DAVIS 2016, that focuses on single objects. 
This is a hard task that allows us to validate the model and to compare with 
the OSMN baseline. As shown in~\autoref{tab:davis2016}, thanks to
the combination of spatio-temporal consistency given by the convLSTM units and
their features modulation, \emph{ReConvNet} outperforms OSMN by $5.4$ points on
the mean IoU ($\mathcal{J}$-mean) metric and is $4$-th in the leaderboard of the 
semi-supervised approaches\footnote{\url{https://davischallenge.org/davis2016/soa_compare.html}}
when comparing on the average between the $\mathcal{J}$ and $\mathcal{F}$ scores.

It is important to highlight that OSVOS~\cite{Caelles2017}, OSVOS-S ~\cite{Man18b} and
onAVOS~\cite{voigtlaender17BMVC} perform online fine-tuning on the first frame
of the video sequence at inference time. Moreover, OSVOS utilizes a boundary
snapping approach, onAVOS makes use of a CRF post-processing step, and OSVOS-S
incorporates instance-aware semantic information from a state-of-the-art
instance segmentation method to further improve the performance.

Most of these methods introduce expensive computation steps at inference time
that are normally not needed when resorting to features modulation. Nothing prevents
though to pair this technique with online fine-tuning or CRF post-processing to 
improve performance. Indeed, with a few steps of fine-tuning at inference time
\emph{ReConvNet} gains $6.9$ points on the $\mathcal{J}$\&$\mathcal{F}$-mean 
placing itself $0.5$ points below onAVOS, which scored 2nd in the public leaderboard.

\subsection{Multiple Objects Segmentation}
\paragraph{DAVIS2017}
The most recent version of DAVIS introduces the challenging task of multiple objects 
segmentation. 
On this dataset \emph{ReConvNet} has been
trained by feeding the visual modulator with one randomly picked object from the scene at 
a time and using the segmentation of the same object in the current frame as target. 
~\autoref{tab:davis2016} shows that \emph{ReConvNet} adapts very well to the
multiobject task outperforming the baseline OSMN by $10.9$ points on the 
$\mathcal{J}$\&$\mathcal{F}$-mean metric. Remarkably, our method also outperforms the state-of-the-art 
OSVOS and onAVOS by $5.4$ and $0.3$ points, respectively, without the need of expensive online
fine-tuning.
Introducing online fine-tuning, the $\mathcal{J}$\&$\mathcal{F}$-mean improves by $4.5$, 
that is $2.2$ points more than OSVOS-S, the current state of the art in the public 
leaderboard on the DAVIS2017 validation set.

\paragraph*{DAVIS Challenge 2018}
We participated to the DAVIS Challenge 2018 retraining on the training set augmented with the validation
set. Our preliminary evaluation on the test-dev set scored $52.7$ and $62.9$ on
$\mathcal{J}$\&$\mathcal{F}$-mean without and with online fine-tuning respectively, 
placing us $8$-th in the test-dev public leaderboard.

On the test-challenge set \emph{ReConvNet} scored $54.5$ $\mathcal{J}$\&$\mathcal{F}$-mean, and $51.8$ 
and $57.2$ $\mathcal{J}$-mean and $\mathcal{F}$-mean, respectively, ranking $10$-th 
in the final DAVIS Challenge 2018 evaluation. This is an encouraging result considering
that no online fine-tuning was employed: by adding gradient steps at inference time it is
reasonable to expect a performance boost similar to the one consistently witnessed in the
previous experiments.

\section{Conclusions and Future work}
We presented \emph{ReConvNet}, a powerful and efficient recurrent convolutional model to perform 
semi-supervised video object segmentation. The model is able to learn spatio-temporal features that 
self-adapt to focus on the object of interest without the need of extra fine-tuning at inference time. 
\emph{ReConvNet} outperforms the baseline by a considerable margin, proving the 
effectiveness of incorporating temporal consistency into the model. Our results reinforce 
the conjecture that features modulation is a valid approach to semi-supervised video object segmentation.
We plan to perform a more in-depth analysis of the interaction between the temporal components
and the features modulation, since we believe it is crucial to better understand the potential of 
the proposed model.

\section*{Acknowledgements}
We thank J\"{u}rgen Schmidhuber and Imanol Schlag for helpful discussions on fast 
weights and Razvan Pascanu for insightful comments on the model. We are also grateful 
to AGS SpA for providing the NVIDIA 1080Ti machine to run all the experiments. 
Finally, our thoughts go to Aaron, Adriana and Michal, for giving the initial thrust to this work.

{\small
\bibliographystyle{ieee}
\bibliography{egbib}

\begin{thebibliography}{10}\itemsep=-1pt

\bibitem{berman2017lov}
M.~Berman, A.~R. Triki, and M.~B. Blaschko.
\newblock The lov\'asz-softmax loss: A tractable surrogate for the optimization
  of the intersection-over-union measure in neural networks.
\newblock {\em arXiv preprint arXiv:1705.08790}, 2017.

\bibitem{Caelles2017}
S.~Caelles, K.-K. Maninis, J.~Pont-Tuset, L.~Leal-Taix\'e, D.~Cremers, and
  L.~{Van Gool}.
\newblock One-shot video object segmentation.
\newblock In {\em CVPR}, 2017.

\bibitem{ha2016hypernetworks}
D.~Ha, A.~Dai, and Q.~V. Le.
\newblock Hypernetworks.
\newblock {\em arXiv preprint arXiv:1609.09106}, 2016.

\bibitem{hariharan2015hypercolumns}
B.~Hariharan, P.~Arbel{\'a}ez, R.~Girshick, and J.~Malik.
\newblock Hypercolumns for object segmentation and fine-grained localization.
\newblock In {\em CVPR}, pages 447--456, 2015.

\bibitem{hochreiter1997long}
S.~Hochreiter and J.~Schmidhuber.
\newblock Long short-term memory.
\newblock {\em Neural computation}, 9(8):1735--1780, 1997.

\bibitem{lin2014microsoft}
T.-Y. Lin, M.~Maire, S.~Belongie, J.~Hays, P.~Perona, D.~Ramanan,
  P.~Doll{\'a}r, and C.~L. Zitnick.
\newblock Microsoft coco: Common objects in context.
\newblock In {\em ECCV}, 2014.

\bibitem{Man18b}
K.-K. Maninis, S.~Caelles, Y.~Chen, J.~Pont-Tuset, L.~Leal-Taix\'e, D.~Cremers,
  and L.~{Van Gool}.
\newblock Video object segmentation without temporal information.
\newblock {\em TPAMI}, 2018.

\bibitem{perez2017film}
E.~Perez, F.~Strub, H.~De~Vries, V.~Dumoulin, and A.~Courville.
\newblock Film: Visual reasoning with a general conditioning layer.
\newblock {\em arXiv preprint arXiv:1709.07871}, 2017.

\bibitem{saxe2013exact}
A.~M. Saxe, J.~L. McClelland, and S.~Ganguli.
\newblock Exact solutions to the nonlinear dynamics of learning in deep linear
  neural networks.
\newblock {\em arXiv preprint arXiv:1312.6120}, 2013.

\bibitem{schmidhuber1992fastweights}
J.~Schmidhuber.
\newblock Learning to control fast-weight memories: An alternative to dynamic
  recurrent networks.
\newblock {\em Neural Computation}, 4(1):131--139, 1992.

\bibitem{simonyan2014very}
K.~Simonyan and A.~Zisserman.
\newblock Very deep convolutional networks for large-scale image recognition.
\newblock {\em arXiv preprint arXiv:1409.1556}, 2014.

\bibitem{dasnet2014}
M.~F. Stollenga, J.~Masci, F.~Gomez, and J.~Schmidhuber.
\newblock Deep networks with internal selective attention through feedback
  connections.
\newblock In {\em Advances in Neural Information Processing Systems 27}. 2014.

\bibitem{Visin2016reseg}
F.~Visin, M.~Ciccone, A.~Romero, K.~Kastner, K.~Cho, Y.~Bengio, M.~Matteucci,
  and A.~Courville.
\newblock Reseg: A recurrent neural network-based model for semantic
  segmentation.
\newblock In {\em CVPR Workshops}, June 2016.

\bibitem{visin_dataset}
F.~Visin and A.~Romero.
\newblock Dataset loaders: a python library to load and preprocess datasets.
\newblock \url{https://github.com/fvisin/dataset_loaders}, 2017.

\bibitem{voigtlaender17BMVC}
P.~Voigtlaender and B.~Leibe.
\newblock Online adaptation of convolutional neural networks for video object
  segmentation.
\newblock In {\em BMVC}, 2017.

\bibitem{xingjian2015clstm}
S.~Xingjian, Z.~Chen, H.~Wang, D.-Y. Yeung, W.-K. Wong, and W.-c. Woo.
\newblock Convolutional lstm network: A machine learning approach for
  precipitation nowcasting.
\newblock In {\em NIPS}, 2015.

\bibitem{Yang2018osmn}
L.~Yang, Y.~Wang, X.~Xiong, J.~Yang, and A.~K. Katsaggelos.
\newblock Efficient video object segmentation via network modulation.
\newblock {\em arXiv preprint arXiv:1802.01218}, 2018.

\end{thebibliography}
}

\end{document}